\documentclass[letterpaper, 10 pt, conference]{ieeeconf}
\IEEEoverridecommandlockouts                              
\usepackage{times}
\usepackage{epsfig}
\usepackage{graphicx}

\usepackage{graphicx}
\usepackage{epsfig}
\usepackage{times}
\usepackage{mathptmx}
\usepackage{cite}
\usepackage{color}
\usepackage{times}

\definecolor{red}{rgb}{1,0,0}
\definecolor{green}{rgb}{0,1,0}
\definecolor{blue}{rgb}{0,0,1}
\definecolor{violet}{rgb}{1,0,1}
\definecolor{cyan}{cmyk}{1,0,0,0}
\definecolor{magenta}{cmyk}{0,1,0,0}
\definecolor{yellow}{cmyk}{0,0,1,0}

\definecolor{white}{rgb}{1,1,1}

\newcommand{\CO}[1]{}

\newcommand{\CommentOut}[1]{}

\newcommand{\noeditage}[1]{#1} \newcommand{\editage}[1]{}

\onecolumn

\begin{document}

\newcommand{\FIG}[3]{
\begin{minipage}[b]{#1cm}
\begin{center}
\includegraphics[width=#1cm]{#2}\\
{\scriptsize #3}
\end{center}
\end{minipage}
}

\newcommand{\FIGU}[3]{
\begin{minipage}[b]{#1cm}
\begin{center}
\includegraphics[width=#1cm,angle=180]{#2}\\
{\scriptsize #3}
\end{center}
\end{minipage}
}

\newcommand{\FIGm}[3]{
\begin{minipage}[b]{#1cm}
\begin{center}
\includegraphics[width=#1cm]{#2}\\
{\scriptsize #3}
\end{center}
\end{minipage}
}

\newcommand{\FIGR}[3]{
\begin{minipage}[b]{#1cm}
\begin{center}
\includegraphics[angle=-90,width=#1cm]{#2}
\\
{\scriptsize #3}
\vspace*{1mm}
\end{center}
\end{minipage}
}

\newcommand{\FIGRpng}[5]{
\begin{minipage}[b]{#1cm}
\begin{center}
\includegraphics[bb=0 0 #4 #5, angle=-90,clip,width=#1cm]{#2}\vspace*{1mm}
\\
{\scriptsize #3}
\vspace*{1mm}
\end{center}
\end{minipage}
}

\newcommand{\FIGCpng}[5]{
\begin{minipage}[b]{#1cm}
\begin{center}
\includegraphics[bb=0 0 #4 #5, angle=90,clip,width=#1cm]{#2}\vspace*{1mm}
\\
{\scriptsize #3}
\vspace*{1mm}
\end{center}
\end{minipage}
}

\newcommand{\FIGpng}[5]{
\begin{minipage}[b]{#1cm}
\begin{center}
\includegraphics[bb=0 0 #4 #5, clip, width=#1cm]{#2}\vspace*{-1mm}\\
{\scriptsize #3}
\vspace*{1mm}
\end{center}
\end{minipage}
}

\newcommand{\FIGtpng}[5]{
\begin{minipage}[t]{#1cm}
\begin{center}
\includegraphics[bb=0 0 #4 #5, clip,width=#1cm]{#2}\vspace*{1mm}
\\
{\scriptsize #3}
\vspace*{1mm}
\end{center}
\end{minipage}
}

\newcommand{\FIGRt}[3]{
\begin{minipage}[t]{#1cm}
\begin{center}
\includegraphics[angle=-90,clip,width=#1cm]{#2}\vspace*{1mm}
\\
{\scriptsize #3}
\vspace*{1mm}
\end{center}
\end{minipage}
}

\newcommand{\FIGRm}[3]{
\begin{minipage}[b]{#1cm}
\begin{center}
\includegraphics[angle=-90,clip,width=#1cm]{#2}\vspace*{0mm}
\\
{\scriptsize #3}
\vspace*{1mm}
\end{center}
\end{minipage}
}

\newcommand{\FIGC}[5]{
\begin{minipage}[b]{#1cm}
\begin{center}
\includegraphics[width=#2cm,height=#3cm]{#4}~$\Longrightarrow$\vspace*{0mm}
\\
{\scriptsize #5}
\vspace*{8mm}
\end{center}
\end{minipage}
}

\newcommand{\FIGf}[3]{
\begin{minipage}[b]{#1cm}
\begin{center}
\fbox{\includegraphics[width=#1cm]{#2}}\vspace*{0.5mm}\\
{\scriptsize #3}
\end{center}
\end{minipage}
}

\newcommand{\figplan}[1]{~}

\newcommand{\SW}[2]{#1}

\title{
\bf 
Open-World Distributed Robot Self-Localization with Transferable Visual Vocabulary and Both Absolute and Relative Features
}

\author{Yoshida Mitsuki ~~~ Yamamoto Ryogo ~~~ Iwata Daiki ~~~ Tanaka Kanji
\thanks{Our work has been supported in part by 
JSPS KAKENHI 
Grant-in-Aid 
for Scientific Research (C) 17K00361, 20K12008, and 23K11270.}
\thanks{The authors are with Graduate School of Engineering,
University of Fukui, Japan. 
{\tt\small mf210369@g.u-fukui.ac.jp,
mf220362@g.u-fukui.ac.jp, tnkknj@u-fukui.ac.jp}}}

\newcommand{\figI}{
\begin{figure}[t]
\begin{center}
\FIG{4}{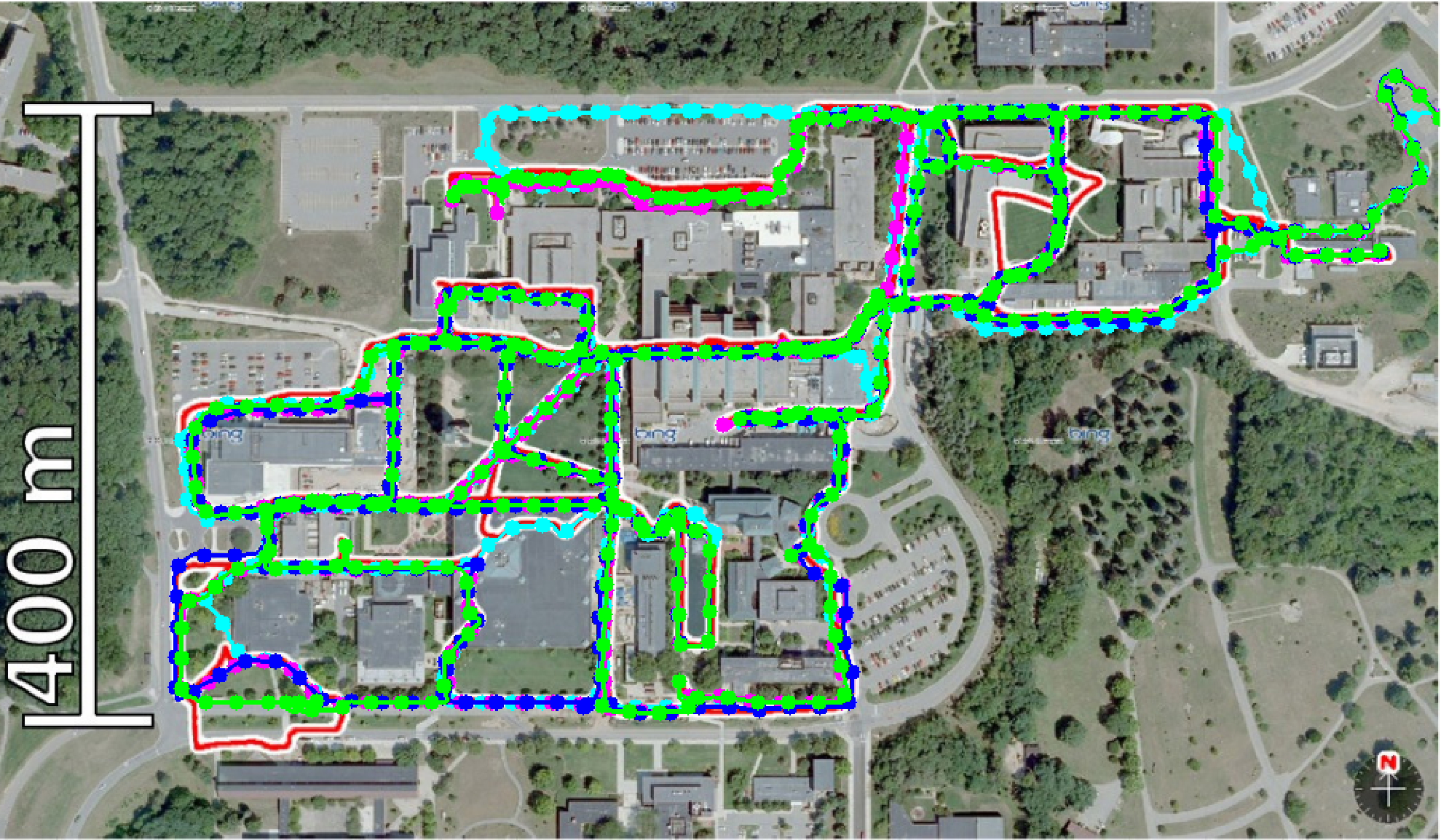}{}%
\caption{Experimental environments.
The trajectories of the four datasets,
``2012/1/22," ``2012/3/31," ``2012/8/4," and ``2012/11/17,"
used in our experiments are visualized 
in green, purple, blue, and light-blue
curves, respectively.
}\label{fig:I}
\end{center}
\end{figure}
}

\newcommand{\figA}{
\begin{figure}[t]
\begin{center}
\FIG{8}{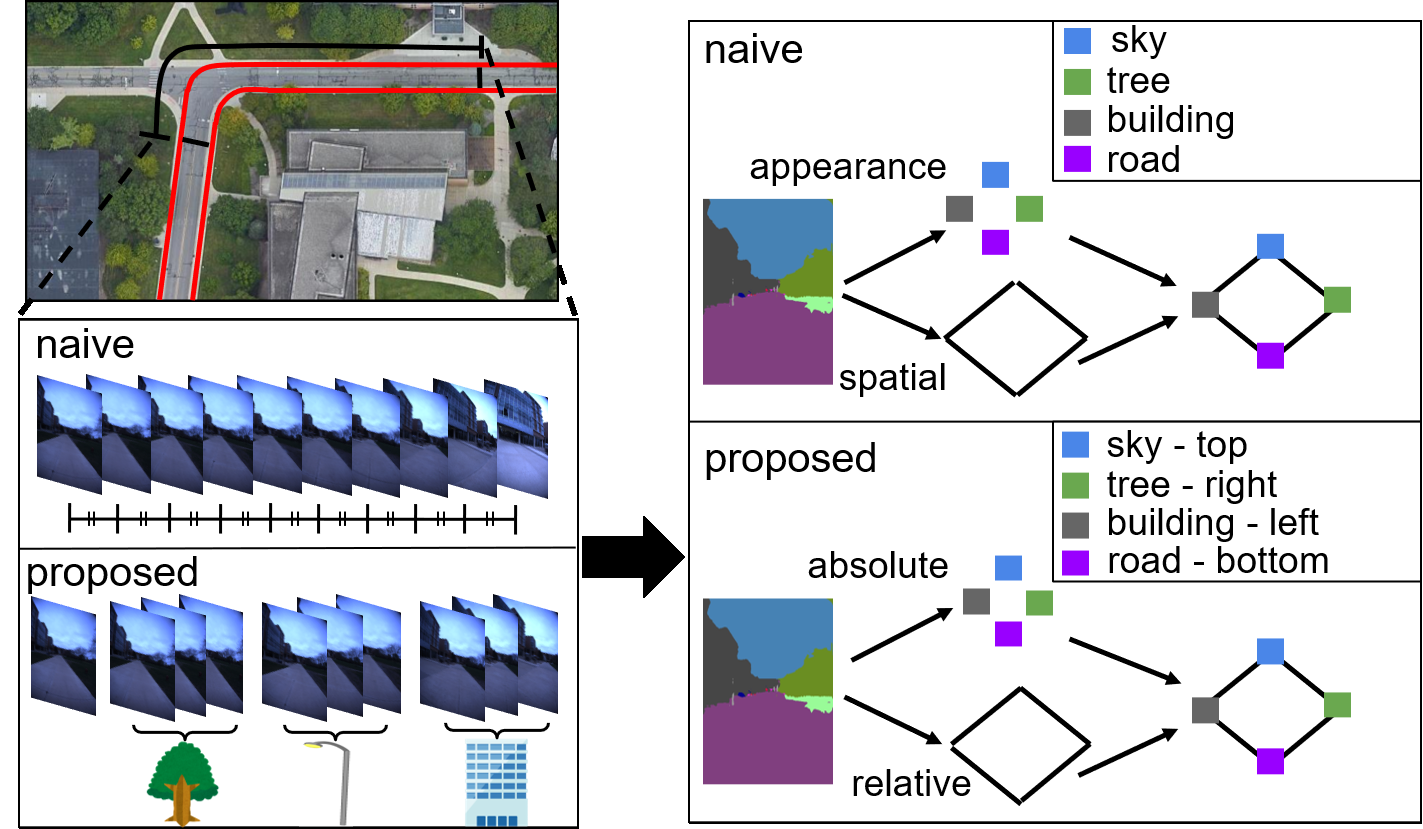}{}
\caption{%
Visual place recognition (VPR) from a novel highly compressive
sequential semantic scene graph (S3G) is considered. To address
the information lost in dimension reduction, the appearance/spatial image
information is mapped to two different features, absolute (node) and relative
(edge) features, which complement each other. Additionally, a new task of
viewpoint planning of the query S3G is enabled by the trained VPR, to
further improve the VPR performance.
}\label{fig:A}
\end{center}
\end{figure}
}

\newcommand{\figB}{
\begin{figure}[t]
\begin{center}
\FIG{4}{figBa.eps}{(a)}
\FIG{4}{figBb.eps}{(b)}\\
\FIG{4}{figBc.eps}{(c)}
\FIG{4}{figBd.eps}{(d)}
\caption{%
\SW{
Different representations of a scene graph. (a) Single-view image graph. (b) Multi-view grid-based segmentation.
(c) Multi-view semantic video clip segmentation. (d) Single-view semantic image segmentation.
}{
}
}\label{fig:B}
\end{center}
\end{figure}
}

\newcommand{\figC}{
\begin{figure}[t]
\begin{center}
\FIG{8}{figC3.eps}{}
\caption{%
\SW{
Bearing-range-semantic (BRS) measurement model. The bearing,  range, and semantics are
observed in an image,
as the location, size, and semantic label, respectively, of an object region. Then, the three-dimensional B-R-S space is quantized to obtain a compact 189-dim 1-hot vector (i.e., an 8-bit descriptor).
}{
}
}\label{fig:C}
\end{center}
\end{figure}
}

\newcommand{\figD}{
\begin{figure}[t]
\begin{center}
\FIG{5}{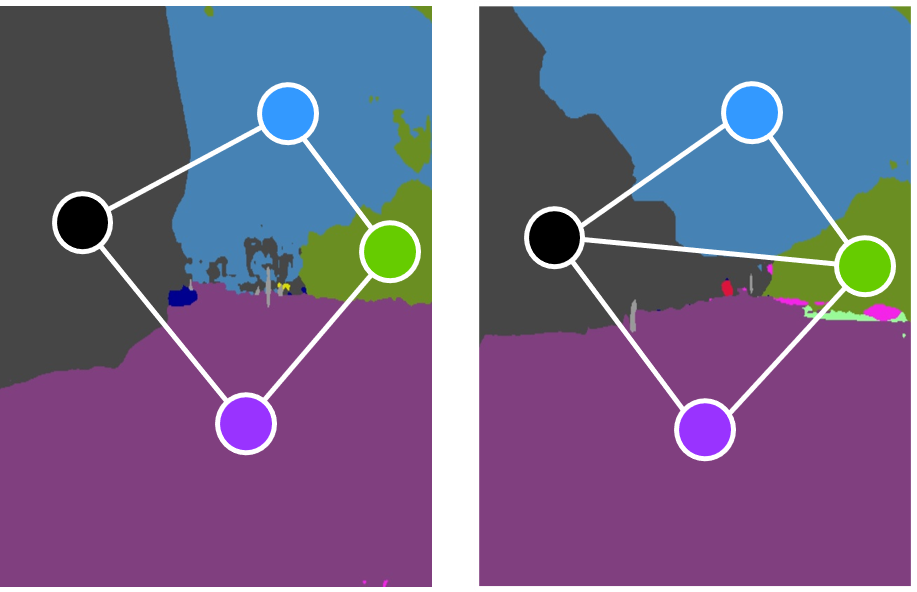}{}
\caption{%
Importance of edges. In highly compressive applications, a naive
strategy of using only absolute features (i.e., nodes) suffer from information
loss during dimension reduction. To address this issue, we exploit the edges
as relative features that complement the absolute features.
}\label{fig:D}
\end{center}
\end{figure}
}

\newcommand{\figE}{
\begin{figure}[t]
\begin{center}
\FIG{8}{figE1.eps}{}
\caption{
\SW{
Algorithm pipeline. In the training/predict stage, input images are converted to an S3G and then fed to the GCN training/test module. In the test stage, the prediction results at each viewpoint are further integrated by a particle filter, to obtain the final estimate.
}{
}
}\label{fig:E}
\end{center}
\end{figure}
}

\newcommand{\figF}{
\begin{figure}[t]
\begin{center}
\FIG{8}{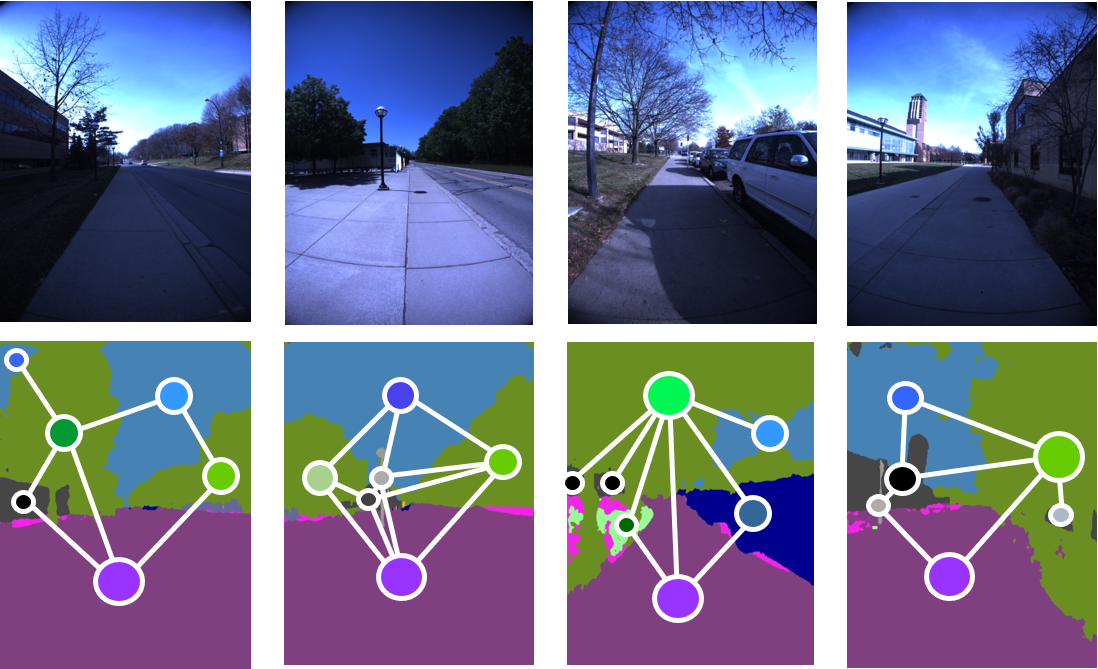}{}
\caption{
\SW{
S2G examples. 
Top: The input image.
Bottom: S2G overlaid on the semantic label image.
}{
}
}\label{fig:F}
\end{center}
\end{figure}
}

\newcommand{\tabA}{

\begin{table}[t]
\begin{center}
\caption{
\SW{
Performance results.
}{
}
}\label{tab:A}

\begin{tabular}{|r|r|r|r|r|r|r|}\hline
\multicolumn{2}{|r|}{}&
\multicolumn{2}{|r|}{w/ region merging}&
\multicolumn{2}{|r|}{w/o region merging}\\ \cline{3-6}
\multicolumn{2}{|r|}{}&
S & BRS & S & BRS \\ \hline
&GCN &11.7 &18.9 &12.0 &19.1 \\ \cline{2-6}
VPR &KNN &5.8 &15.6 &6.3 &12.9   \\ \cline{2-6}
&NBNN &1.3 &3.4 &1.4 &3.5 \\ \hline
&GCN & - &30.6 & - & -   \\ \cline{2-6}
VPR+VP &KNN & - & - & - &  -  \\ \cline{2-6}
&NBNN &- & -&- & -   \\ \hline
\end{tabular}
\end{center}
\end{table}

}

\newcommand{\tabB}{

\begin{table}[t]
\caption{%
Ablation studies.
}\label{tab:B}
\scalebox{1.0}{
\begin{tabular}{|rr|r|r|r|r|r|}\hline
&&(SP, SU)&(SU, AU)&(AU, WI)&(WI, SP) \\ \hline
Fig. \ref{fig:B}d &(1-hot) &23.8 &14.8 &15.3 &21.7 \\ \hline
Fig. \ref{fig:B}a &&21.4 &13.9 &13.7 &20.1   \\ \hline
Fig. \ref{fig:B}d &(3-hot) &20.1 &12.2 &13.7 &19.2   \\ \hline
\end{tabular}
}
\end{table}

}

\newcommand{\hs}{\hspace*{1.7cm}}

\newcommand{\figG}{

\begin{figure}[t]
\begin{center}
\FIG{6}{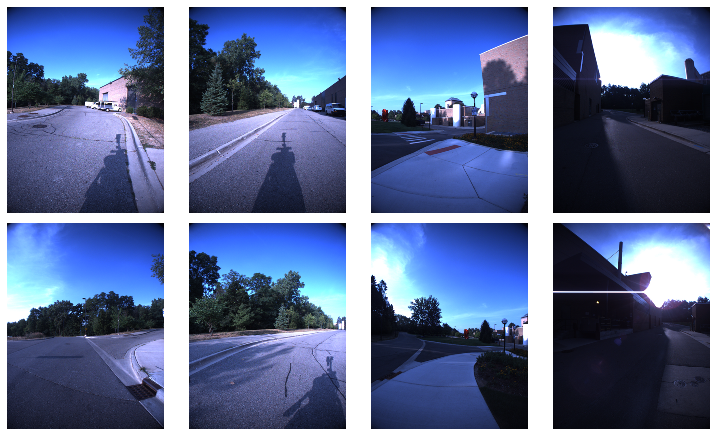}{}\vspace*{3mm}\\
\scriptsize
\hspace*{8mm}(a)\hs (b)\hs (c)\hs (d)\hs
\caption{%
NBV planning results. In each figure, the bottom and top panels show the view image before and after planned movements, respectively.
}\label{fig:G}
\end{center}
\end{figure}

}

\maketitle

\author{}

\begin{abstract}
Visual robot self-localization is a fundamental problem in visual robot navigation and has been studied across various problem settings, including monocular and sequential localization. However, many existing studies focus primarily on single-robot scenarios, with limited exploration into general settings involving diverse robots connected through wireless networks with constrained communication capacities, such as open-world distributed robot systems. In particular, issues related to the transfer and sharing of key knowledge, such as visual descriptions and visual vocabulary, between robots have been largely neglected.
This work introduces a new self-localization framework designed for open-world distributed robot systems that maintains state-of-the-art performance while offering two key advantages: (1) it employs an unsupervised visual vocabulary model that maps to multimodal, lightweight, and transferable visual features, and (2) the visual vocabulary itself is a lightweight and communication-friendly model. Although the primary focus is on encoding monocular view images, the framework can be easily extended to sequential localization applications. By utilizing complementary similarity-preserving features---both absolute and relative---the framework meets the requirements for being unsupervised, multimodal, lightweight, and transferable. All features are learned and recognized using a lightweight graph neural network and scene graph. The effectiveness of the proposed method is validated in both passive and active self-localization scenarios.
\end{abstract}

\section{%
Introduction
}

Visual robot self-localization is a fundamental problem in visual robot navigation and has been studied across various problem settings, including monocular and sequential localization. However, many existing studies focus primarily on single-robot scenarios, with limited exploration into general settings involving diverse robots connected through wireless networks with constrained communication capacities, such as open-world distributed robot systems. In particular, issues related to the transfer and sharing of key knowledge, such as visual descriptions and visual vocabulary, between robots have been largely neglected.

Image sequence-based self-localization has received considerable attention as a highly compressive and discriminative approach to long-term visual place recognition (VPR) across domains (e.g., weather, time of day, and season). Highly-compressive scene representation is essential for robots to perform long-term and large-scale VPR tasks in virtual training and real deployment environments. Given a long-term image sequence $I_1^{map}, \cdots, I_T^{map}$ covering the robot workspace in a past domain ("map"), the sequence-based self-localization aims to determine the most matched sub-sequence for a short-term query sequence $I_{t_1}^{query}, \cdots, I_{t_N}^{query}$ $(|t_{N}-t_{1}|\ll T)$ in a new unseen domain \cite{cs1}. An advantage of this sequence-based self-localization approach is its good balance between discriminative power and compactness. Even in highly compressive applications (e.g., 4-bit image descriptor \cite{yan2019global}) where typical single-view self-localization approaches struggle, the sequence-based self-localization approach may maintain high performance due to multi-view information fusion. Moreover, such a map can be flexibly reorganized into a local map $I_a, \cdots, I_b$ with arbitrary start $a$ and goal $b$ viewpoints ($|b-a|<T$), enabling efficient reuse of storage in applications such as multi-robot multi-session knowledge sharing \cite{sharing}.

In this study, we explore the sequence-based self-localization problem from a novel perspective of sequential semantic scene graph (S3G), as shown in Fig. \ref{fig:A}. The motivation for this approach is threefold. First, the semantic scene graph (S2G) $x^G=\langle x^N, x^E \rangle$, where nodes $x^N$ represent semantic regions (e.g., object regions) and edges $x^E$ represent relationships between nodes, is one of the most compact scene descriptors in computer vision \cite{SceneGraph}. Second, recently-developed deep graph convolutional neural networks (GCNs) can serve as a powerful visual place classifier (VPC) $y=f_{VPC}(x)$ that takes a scene graph $x$ as input and predicts the place class $y$ \cite{icra21takeda}. Third, such a well-trained VPC provides useful information for viewpoint planning (VP) of the query sequence $I_{t_1}^{query}, \cdots, I_{t_N}^{query}$, which further improves the VPR performance. The remaining challenge is designing an effective similarity-preserving mapping $x^G=f_{I2G}(I)$ from an input image $I$ to a scene graph $x^G$.

Here, we propose to exploit a pair of similarity-preserving mappings: image-to-nodes $x^N=f_{I2N}(I)$ and image-to-edges $x^E=f_{I2E}(I)$, which provide two forms of appearance/spatial image features. In earlier approaches, appearance and spatial image features are separately mapped to nodes and edges, respectively (i.e., appearance-to-nodes, and spatial-to-edges) \cite{AppearanceNodesSpatialEdges}. However, in our highly compressive application, the amount of information lost in dimension reduction is significant, which can frequently lead to recognition errors in self-localization. In the proposed approach, nodes $x^N$ and edges $x^E$ act as {\it absolute} (e.g., size and brightness) and {\it relative} (e.g., larger and brighter) features that complement each other, thereby further improving overall VPR performance.

Our contributions are as follows:
\begin{enumerate}
\item
We propose an unsupervised visual vocabulary model that provides multimodal, lightweight, and transferable features for both monocular and sequential localization applications.
\item
We designed the visual vocabulary model to be lightweight and communication-friendly, with features learned and recognized through a lightweight graph neural network and scene graph.
\item
We develop a prototype system implementing the proposed framework for simultaneous visual place recognition (VPR) and viewpoint planning. 
\end{enumerate}
The effectiveness of this method is validated in both passive and active self-localization scenarios using the public NCLT dataset \cite{NCLT}.

\noeditage{
\figA
}

\section{%
Related Work
}

Highly compressive scene descriptors have been extensively studied in long-term, large-scale Visual Place Recognition (VPR) applications. In single-view VPR applications, vector quantization techniques, such as bag-of-words \cite{sivic2003video}, and compact binary codes based on either local \cite{sivic2003video} or global features \cite{GIST} have been explored. More recently, methods have been developed to pack local image features into compact binary codes to further reduce the bits per image \cite{pbow}. Compared to these single-view approaches, sequence-based self-localization requires fewer bits per image due to multi-view information fusion. Our approach combines the advantages of robust multi-view self-localization and discriminative scene graph descriptors.

The field of sequence-based self-localization was pioneered by the seminal work of SeqSLAM \cite{topo2}, the first cross-domain self-localization system. Subsequent research has focused on improving descriptive power \cite{seqslamDescriptive} and inference algorithms \cite{seqslamInference}. While most existing works concentrate on enhancing VPR accuracy and reducing execution time, our work emphasizes graph neural networks and scene graph compression, along with a new task of viewpoint planning. This is particularly effective for autonomous navigation scenarios, where the robot must adjust its viewpoints to minimize energy consumption during movement.

Recent research has shown increased interest in graph-based scene representation for self-localization \cite{ref10,9665885,7780860}. For instance, \cite{ref10} decouples a scene into subimages via semantic segmentation and connects these subimages with object-level edges, demonstrating good VPR performance.

Earlier approaches often describe nodes with high-dimensional visual features. However, this method is costly in highly compressive applications \cite{GraphVprHighDimNode}. Our approach utilizes lightweight feature category IDs instead of expensive high-dimensional features.

Graph-based VPR has been formulated as a graph matching task \cite{GraphVprGraphMatching}. However, the cost of graph matching algorithms increases rapidly with environment size. Our approach formulates graph-based VPR as a graph classification task, inheriting the benefits of classification tasks, such as flexibility in defining place classes \cite{ref6}, a compressed classifier model \cite{CNNcompression}, and high classification speed \cite{distil}.

Our active vision approach, based on viewpoint planning, is inspired by recent advancements in active self-localization. The work of \cite{burgard1997active} extends the Markov localization framework for action planning, while \cite{feder1999adaptive} presents an appearance-based active observer for micro-aerial vehicles. The method in \cite{chaplot2018active} addresses active self-localization using a learned policy model with deep neural networks. Additionally, \cite{deepactivelocalization} completely learns the policy model, perceptual model, and likelihood model, and \cite{chaplot2020learning} investigates a neural network-based active SLAM framework. However, these studies assume in-domain scenarios where appearance changes between training and testing domains are minimal. Our approach tackles this issue from a novel perspective of domain-invariant VPR. To the best of our knowledge, no prior work has applied Graph Convolutional Networks (GCNs) in this context.

\section{%
Approach
}\label{sec:approach}

The system includes the 
(offline) training module
and
(online) test module.
In addition,
a scene graph descriptor sub-module
is 
employed by both the training and test modules.
These modules are detailed in the following subsections.

\subsection{%
Edge: Relative Feature
}\label{sec:relative}

We considered several possible approaches for constructing graph edges from an input query/map image sequence. 

The first approach views each image frame in the sequence as a graph node and connects neighboring image nodes with edges \cite{icra21takeda}. The second approach applies pre-defined partitioning, such as grid-based partitioning, to obtain spatio-temporal node regions. The third approach uses semantic video segmentation techniques to segment an input image sequence (video clip) into spatio-temporal region nodes. The fourth approach employs image segmentation techniques to segment images into image region nodes.

In this study, we chose the image scene graph with image region nodes (i.e., the fourth approach) for the following reasons. Our earlier system on Graph Convolutional Network (GCN)-based VPR \cite{icra21takeda} used predefined regions for each image frame (e.g., ``center," ``left," ``right"). However, this graph structure did not accurately reflect the scene layout and thus provided limited information. We used this method as a baseline in our experiments.

The second and third approaches are prohibitive for our application; they require the start/goal endpoints of individual sequences to be defined during the training stage, which cannot be adjusted on the fly during the testing stage. In contrast, the fourth approach is both flexible and informative. Unlike the second and third approaches, it allows the map-user robot to dynamically control the endpoints during the testing stage. Furthermore, unlike the first approach, the graph edges in this method accurately reflect the scene layout, resulting in more informative scene graphs.

The procedure for graph construction is as follows. First, semantic labels are assigned to pixels using DeepLab v3+ \cite{chen2018encoder}, pretrained on the Cityscapes dataset. Regions smaller than 100 pixels are regarded as noise and removed. Connected regions with the same semantic labels are identified using a flood-fill algorithm \cite{FloodFillAlgorithm}, with each region assigned a unique region ID. Each region is then connected to its neighboring regions by edges. Finally, an image scene graph with image region nodes is obtained.

\subsection{%
Node: Absolute Feature
}

Image region descriptors (i.e., node descriptors in our case) have been extensively studied. Existing descriptors can generally be categorized into two groups: local features \cite{SIFT} and global features \cite{GIST}. Most descriptors are either high-dimensional feature vectors \cite{SIFT} or unordered collections of vector quantized features known as "bag-of-words" \cite{sivic2003video}. However, these descriptors incur significant space costs that increase with the number and dimensionality of the original feature vectors, often requiring hundreds or thousands of bytes per scene. This poses a challenge in highly compressive applications such as ours.

In the field of image retrieval, semantic labels (e.g., object category IDs) have been utilized as extremely compact descriptors for indexing and retrieving images \cite{CategoryBasedRetrieval}. 

A key difference between our robotics application and image retrieval applications is the importance of surrounding object semantics and their spatial information, such as bearing ("B") and range ("R"), in visual place recognition (VPR). In robotic SLAM, special interest has been given to various types of spatial information, including range-bearing SLAM \cite{BearingRangeSlam}, range-only SLAM \cite{RangeOnlySlam}, and bearing-only SLAM \cite{BearingOnlySlam}.

Based on these considerations, our VPR task was formulated using bearing-range-semantic (BRS) measurements.

Specifically, our approach involves re-categorizing the semantic labels output by a semantic segmentation network \cite{chen2018encoder} into seven distinct semantic category IDs: "sky," "tree," "building," "pole," "road," "traffic sign," and "the others." These correspond to the original labels \{``sky"\}, \{``vegetation"\}, \{``building"\}, \{``pole"\}, \{``road," ``sidewalk"\}, \{``traffic-light," ``traffic-sign"\}, and \{``person," ``rider," ``car," ``truck," ``bus," ``train," ``motorcycle," ``bicycle," ``wall," ``fence," ``terrain"\}. The center location of each region was quantized into nine "bearing" category IDs using a 3$\times$3 regular grid. The region size was quantized into three "range" category IDs: "short distance" (larger than 150 K pixels), "medium distance" (50 K-150 K pixels), and "long distance" (smaller than 50 K pixels) for a 616$\times$808 image. Finally, these semantic, bearing, and range category IDs are combined to form a 189-dimensional one-hot vector as the node descriptor.

\noeditage{
\figD
}

At first glance, the use of edges as additional features might appear unnecessary, given that both appearance and spatial features are already incorporated in the node descriptors. However, the objectives of edge descriptors are quite different. Specifically, edge descriptors are more suited for capturing relative features (e.g., position relationships) rather than absolute features (e.g., position). 

Nodes and edges can act as complementary error detection codes \cite{GeneralErrorDetectionCode}, enhancing each other's effectiveness. For example, as illustrated in Figure \ref{fig:D}, edge features play a crucial role in discriminating between nearly duplicate Scene Graph (S2G) pairs, a task that cannot be achieved with node descriptors alone.

\subsection{%
Region Merging 
}\label{sec:merging}

Most existing techniques for semantic segmentation trade domain-invariance for accuracy. Their objective function and performance metrics prioritize pixel-level precision and recall, aiming to detect semantically coherent regions with the highest accuracy. Consequently, the graph topology produced by these methods is often adversely affected by factors such as viewpoint drift and occlusions. This poses a challenge for our Visual Place Recognition (VPR) application, as the graph topology serves as a primary cue that significantly influences the performance of Graph Convolutional Networks (GCNs) \cite{icra21takeda}.

To mitigate this issue, we adopted a region merging technique inspired by recent work \cite{cvpr19_merging_small}. Specifically, we eliminate small regions with areas smaller than 1,000 pixels (for a 616$\times$808 image). This simple technique has proven to be highly effective in enhancing VPR performance, as demonstrated through ablation studies presented in Section \ref{sec:exp}.

\subsection{%
Self-localization from S2G
}

Self-localization using Semantic Scene Graphs (S2Gs) is effectively addressed by Graph Convolutional Network (GCN)-based Visual Place Classifiers (VPCs). This approach takes a single-view image and predicts the place class. GCNs are a prominent type of graph neural network and have been successfully applied in various domains, including web-scale recommender systems \cite{gcn2} and chemical reactivity prediction \cite{gcn1}. In our recent study \cite{icra21takeda}, a GCN was trained as an image-sequence classifier and achieved state-of-the-art performance in Visual Place Recognition (VPR).

For defining place classes, we follow the grid-based partitioning method described in \cite{ref1}. In the experimental environment, this method results in a 10$\times$10 grid, yielding a total of 100 place classes.

In this study, we trained a GCN using S2Gs as the training data. The graph convolution operation processes each node $v_i$ in the following manner. Initially, the node receives messages from its neighboring nodes connected by edges. These messages are aggregated by summing them via the SUM function. The aggregated result is then passed through a single-layer fully connected neural network, followed by a nonlinear transformation to generate a new feature vector. We used the rectified linear unit (ReLU) operation as the nonlinear transformation in this study. This process is applied to all nodes in the graph during each iteration, producing a new graph with updated node features while maintaining the original graph's structure.

The iterative process is repeated $L$ times, where $L$ denotes the number of GCN layers. After averaging the graph node information obtained from these iterations, the probability value vector for the graph's prediction is derived by applying a fully connected layer and the softmax function. For implementation, we utilized the Deep Graph Library (DGL) \cite{wang2019dgl} on the PyTorch backend.

\subsection{%
Self-localization from S3G
}

The sequence-based self-localization system aims to estimate the robot's location from a sequence of Semantic Scene Graphs (S2Gs), collectively referred to as Sequential Semantic Scene Graphs (S3Gs). This system is composed of two primary modules: information fusion and viewpoint planning.

The information fusion module utilizes a particle filter (PF) to integrate the history of perceptual-action measurements $I_{t_1}^{query}, \cdots, I_{t_n}^{query}$ at each viewpoint $t_n$ of the query image sequence into the current belief of the robot's location \cite{MCL}. In this context, an action corresponds to a forward movement along the viewpoint trajectory, while a perception corresponds to a class-specific probability density vector (PDV) output by the GCN.

Our approach differs from traditional PF-based self-localization systems in that both perceptual measurements and beliefs are represented using class-specific reciprocal rank vectors. Specifically, the $i$-th element of the PDV is the reciprocal rank value of the $i$-th class, computed as the reciprocal of the rank of the class in a list sorted in descending order of probability values \cite{icra21takeda}. During the measurement update, each particle is assigned to a place class, and its weight is updated by adding the reciprocal rank value of the class corresponding to the particle's hypothesized location.

Viewpoint planning is formulated as a reinforcement learning (RL) problem, where a learning agent interacts with a stochastic environment. This interaction is modeled as a discrete-time discounted Markov decision process (MDP), represented as a quintuple $(S, A, P, R, \gamma)$. Here, $S$ and $A$ denote the sets of states and actions, respectively, $P$ represents the state transition distribution, $R$ denotes the reward function, and $\gamma \in (0,1)$ is the discount factor ($\gamma = 0.9$).

The probability distribution over the next state and the immediate reward of performing action $a$ in state $s$ are denoted by $P(\cdot | s, a)$ and $r(s, a)$, respectively. The state $s$ is defined as the class-specific reciprocal rank vector \cite{civemsa2021kanji}, and action $a$ corresponds to a forward movement along the route. In our experiments, the action candidate set was $A = \{1, 2, \cdots, 10\}$ meters. Each training/test episode consists of 10 sequential actions.

To address the curse of dimensionality in RL, we employed the nearest neighbor approximation of Q-learning (NNQL) \cite{nnql} with $k=4$. The immediate reward was provided at the final viewpoint of each training episode, calculated as the reciprocal rank value of the ground truth viewpoint.

\section{
Experiments
}\label{sec:exp}

The proposed method was evaluated in a cross-season self-localization scenario to validate whether the GCN-based Visual Place Recognition (VPR) method could enhance performance in both passive and active VPR scenarios.

The public NCLT dataset \cite{NCLT} was employed for the experiments, as shown in Figure \ref{fig:I}. This dataset, collected using a Segway vehicle at the University of Michigan North Campus, encompasses long-term autonomy with both indoor and outdoor environments. The dataset presents various geometric changes (e.g., object placement, pedestrian movements, car parking/stopping) and photometric changes (e.g., lighting variations, shadows, occlusions).

In particular, we considered challenging cross-season self-localization scenarios, where the self-localization system is trained and tested across different seasons. We created four (training, test) season pairings from the dataset: ``2012/1/22 (WI)," ``2012/3/31 (SP)," ``2012/8/4 (SU)," and ``2012/11/17 (AU)". The pairings were (WI, SP), (SP, SU), (SU, AU), and (AU, WI). Additionally, an extra season, ``2012/5/11 (EX)," was used for training the VPR system. The VPR was trained once using the EX season data and the learned parameters were applied across all the training-testing season pairs.

Figure \ref{fig:F} illustrates scene graphs obtained using the proposed scene graph construction procedure. Notably, domain-invariant elements, such as buildings and roads, were predominantly selected as image region nodes.

We evaluated three different VPR methods: GCN, naive Bayes nearest neighbor (NBNN), and k-nearest neighbor (kNN). All methods utilized an image sequence represented as an S3G. The GCN is the proposed method detailed in Section \ref{sec:approach}. Given that prior studies on highly compressive S3G applications are limited, our investigation focused on developing and evaluating various possible ablations of this method. 

NBNN \cite{nbnn} is well-established for measuring dissimilarities between feature sets, demonstrating high performance in previous studies \cite{our_previous_nbnn}. NBNN calculates the L2 distance from the nearest-neighbor map feature to each query feature and averages this over all query features to yield the NBNN dissimilarity value.

kNN, a traditional non-parametric classification method, classifies based on the nearest-neighbor training samples in the feature space. The class labels most frequently assigned to the k nearest neighbors (i.e., minimum L2 norm) are used as the classification result. In this approach, an image is described by a 189-dimensional histogram vector that aggregates all node features belonging to the image.

\figI

\figF

The Visual Place Recognition (VPR) performance was evaluated in terms of top-1 accuracy. The evaluation procedure was conducted as follows:

\begin{enumerate}
    \item The VPR performance was assessed at all viewpoints of the query sequence, not limited to the final viewpoint.
    \item For each viewpoint, the top-1 accuracy was computed based on the latest output from the particle filter. Specifically, the class with the highest belief value was compared to the ground-truth class to determine if they matched.
\end{enumerate}

\figG

\tabA

\section{Results and Discussion}

\subsection{Behavior of the Robot}

Figure \ref{fig:G} illustrates examples of views before and after planned Next-Best-View (NBV) actions. The behavior of the robot was intuitively convincing. Prior to the move, the scene was either non-salient, consisting only of the sky, road, and trees (Figure \ref{fig:G} a, b, c), or the field of view was very narrow due to occlusions (Figure \ref{fig:G} d). After the move, landmark objects came into view (Figure \ref{fig:G} a, c) or additional landmarks appeared (Figure \ref{fig:G} b, d). Such behaviors are intuitively appropriate and effective for seeking and tracking landmarks, similar to how humans look for familiar landmarks when lost. Our approach enables the robot to learn such appropriate step sizes from available visual experience.

\subsection{Ablation Study}

An ablation study was conducted to observe the effects of individual components, including the relative edge feature (Section \ref{sec:relative}) and the region merging technique (Section \ref{sec:merging}). Table \ref{tab:A} presents the results for all combinations of the proposed and baseline methods with and without viewpoint planning (VP). Notably, the proposed method demonstrated superior performance compared to other methods. The region merging technique (Section \ref{sec:merging}) contributed to reducing the number of nodes while retaining VPR performance. Specifically, the number of nodes was reduced from an average of 19.8 to 7.2 per S2G, which resulted in a reduction in computation time for VPR from 0.82 ms to 0.16 ms. Particularly, the use of the edge feature and the NBV module often significantly enhanced VPR performance.

\subsection{Space Costs}

Finally, we investigated space costs. The average number of nodes per S2G was 7.2. Each node descriptor consumed 8 bits. The space costs for nodes and edges were 57.8 bits and 12.5 bits per S2G, respectively. These costs are significantly lower than those of typical methods based on high-dimensional vectorial features and compact variants such as bag-of-words. Notably, the current descriptors were not compressed, suggesting that further compression may be achievable.

\section{
Conclusions
}

Existing studies on visual robot self-localization, which cover monocular and sequential contexts, often concentrate on single-robot scenarios and overlook the challenges of transferring and sharing critical visual knowledge between robots in open-world distributed systems.
In this study, we proposed a novel, highly compressive Visual Place Recognition (VPR) framework for sequence-based self-localization using Scene-to-Graph (S3G) representations. To address the significant information loss resulting from dimensionality reduction, we introduced a pair of similarity-preserving mappings?image-to-nodes and image-to-edges?that complement each other. Additionally, we applied the S3G framework to the viewpoint-planning task, enabling dynamic control of viewpoints to further enhance performance. We validated the effectiveness of the proposed method through experiments using the public NCLT dataset, including performance comparisons and ablation studies, demonstrating its effectiveness in both passive and active self-localization scenarios.

\bibliographystyle{IEEEtran}
\bibliography{ieee}

\end{document}